\documentclass[10pt,twocolumn,letterpaper]{article}

\usepackage{cvpr}
\usepackage{times}
\usepackage{epsfig}
\usepackage{graphicx}
\usepackage{amsmath}
\usepackage{amssymb}
\usepackage{nicefrac}
\usepackage{fixltx2e} 
\usepackage{leftidx}

\usepackage{algorithm}
\usepackage{algpseudocode}
\usepackage{booktabs}

\usepackage{tikz}
\usetikzlibrary{shapes.geometric}
\usepackage{rotating}
\usetikzlibrary{positioning}

\usepackage{pifont}
\usepackage{multirow}

\newcommand{\finalcopy}{\cvprfinalcopy}
\usepackage{tikz}
\usetikzlibrary{arrows,shapes,calc,matrix,fit,backgrounds}
\usepackage{pgfplots}
\usepackage{pgfplotstable}
\pgfplotsset{compat=1.9}

\usepackage{xstring}

\usepgfplotslibrary{external}
\newcommand{\extfig}[2]{\tikzsetnextfilename{fig/extern/#1}{#2}}
\newcommand{\extdata}[1]{\input{#1}}
\IfBeginWith*{\jobname}{fig/extern/}{\finalcopy}{}

\newcommand{\leg}[1]{\addlegendentry{#1}}

\tikzset{every mark/.append style={solid}}
\pgfplotsset{
	grid=both, width=\columnwidth, try min ticks=5,
	every axis/.append style={font=\scriptsize},
	every axis plot/.append style={thick,mark=none,mark size=1.2,tension=0.18},
	legend cell align=left, legend style={fill opacity=0.8},
}

\pgfplotsset{
	dash/.style={mark=o,dashed,opacity=0.7},
	dott/.style={mark=o,dotted,opacity=0.7},
}

\usepackage[pagebackref=true,breaklinks=true,colorlinks,bookmarks=false]{hyperref}

\cvprfinalcopy 

\def\cvprPaperID{2572} 

\ifcvprfinal\fi

\begin{document}

\title{Label Propagation for Deep Semi-supervised Learning}

\author{
Ahmet Iscen$^1$ \ \ \ \ Giorgos Tolias$^1$\ \ \ \ Yannis Avrithis$^2$\ \ \ \ Ond{\v r}ej Chum$^{1}$\\
{\fontsize{11}{13}\selectfont$^1$VRG, FEE, Czech Technical University in Prague\ \ \ \ \ \ $^2$Univ Rennes, Inria, CNRS, IRISA}\\
}

\newcommand{\nn}[1]{\ensuremath{\text{NN}_{#1}}\xspace}

\newcommand{\prm}[1]{_{#1}}
\newcommand{\dime}[1]{(#1)}

\def\l1{\ensuremath{\ell_1}\xspace}
\def\l2{\ensuremath{\ell_2}\xspace}

\newcommand*\OK{\ding{51}}

\newenvironment{narrow}[1][1pt]
	{\setlength{\tabcolsep}{#1}}
	{\setlength{\tabcolsep}{6pt}}

\newcommand{\comment} [1]{{\color{orange} \Comment     #1}} 


\newcommand{\head}[1]{{\smallskip\noindent\bf #1}}
\newcommand{\eq}[1]{(\ref{eq:#1})\xspace}

\newcommand{\red}[1]{{\color{red}{#1}}}
\newcommand{\blue}[1]{{\color{blue}{#1}}}
\newcommand{\green}[1]{{\color{green}{#1}}}
\newcommand{\gray}[1]{{\color{gray}{#1}}}


\newcommand{\tran}{^\top}
\newcommand{\mtran}{^{-\top}}
\newcommand{\zcol}{\mathbf{0}}
\newcommand{\zrow}{\zcol\tran}

\newcommand{\ind}{\mathbbm{1}}
\newcommand{\expect}{\mathbb{E}}
\newcommand{\nat}{\mathbb{N}}
\newcommand{\zahl}{\mathbb{Z}}
\newcommand{\real}{\mathbb{R}}
\newcommand{\proj}{\mathbb{P}}
\newcommand{\prob}{\mathbf{Pr}}

\newcommand{\mif}{\textrm{if }}
\newcommand{\other}{\textrm{otherwise}}
\newcommand{\minimize}{\textrm{minimize }}
\newcommand{\maximize}{\textrm{maximize }}
\newcommand{\st}{\textrm{subject to }}

\newcommand{\id}{\operatorname{id}}
\newcommand{\const}{\operatorname{const}}
\newcommand{\sgn}{\operatorname{sgn}}
\newcommand{\var}{\operatorname{Var}}
\newcommand{\mean}{\operatorname{mean}}
\newcommand{\trace}{\operatorname{tr}}
\newcommand{\diag}{\operatorname{diag}}
\newcommand{\vect}{\operatorname{vec}}
\newcommand{\cov}{\operatorname{cov}}

\newcommand{\softmax}{\operatorname{softmax}}
\newcommand{\clip}{\operatorname{clip}}

\newcommand{\defn}{\mathrel{:=}}
\newcommand{\peq}{\mathrel{+\!=}}
\newcommand{\meq}{\mathrel{-\!=}}

\newcommand{\floor}[1]{\left\lfloor{#1}\right\rfloor}
\newcommand{\ceil}[1]{\left\lceil{#1}\right\rceil}
\newcommand{\inner}[1]{\left\langle{#1}\right\rangle}
\newcommand{\norm}[1]{\left\|{#1}\right\|}
\newcommand{\frob}[1]{\norm{#1}_F}
\newcommand{\card}[1]{\left|{#1}\right|\xspace}
\newcommand{\diff}{\mathrm{d}}
\newcommand{\der}[3][]{\frac{d^{#1}#2}{d#3^{#1}}}
\newcommand{\pder}[3][]{\frac{\partial^{#1}{#2}}{\partial{#3^{#1}}}}
\newcommand{\ipder}[3][]{\partial^{#1}{#2}/\partial{#3^{#1}}}
\newcommand{\dder}[3]{\frac{\partial^2{#1}}{\partial{#2}\partial{#3}}}

\newcommand{\wb}[1]{\overline{#1}}
\newcommand{\wt}[1]{\widetilde{#1}}

\def\xssp{\hspace{1pt}}
\def\ssp{\hspace{3pt}}
\def\msp{\hspace{5pt}}
\def\lsp{\hspace{12pt}}

\newcommand{\cA}{\mathcal{A}}
\newcommand{\cB}{\mathcal{B}}
\newcommand{\cC}{\mathcal{C}}
\newcommand{\cD}{\mathcal{D}}
\newcommand{\cE}{\mathcal{E}}
\newcommand{\cF}{\mathcal{F}}
\newcommand{\cG}{\mathcal{G}}
\newcommand{\cH}{\mathcal{H}}
\newcommand{\cI}{\mathcal{I}}
\newcommand{\cJ}{\mathcal{J}}
\newcommand{\cK}{\mathcal{K}}
\newcommand{\cL}{\mathcal{L}}
\newcommand{\cM}{\mathcal{M}}
\newcommand{\cN}{\mathcal{N}}
\newcommand{\cO}{\mathcal{O}}
\newcommand{\cP}{\mathcal{P}}
\newcommand{\cQ}{\mathcal{Q}}
\newcommand{\cR}{\mathcal{R}}
\newcommand{\cS}{\mathcal{S}}
\newcommand{\cT}{\mathcal{T}}
\newcommand{\cU}{\mathcal{U}}
\newcommand{\cV}{\mathcal{V}}
\newcommand{\cW}{\mathcal{W}}
\newcommand{\cX}{\mathcal{X}}
\newcommand{\cY}{\mathcal{Y}}
\newcommand{\cZ}{\mathcal{Z}}

\newcommand{\vA}{\mathbf{A}}
\newcommand{\vB}{\mathbf{B}}
\newcommand{\vC}{\mathbf{C}}
\newcommand{\vD}{\mathbf{D}}
\newcommand{\vE}{\mathbf{E}}
\newcommand{\vF}{\mathbf{F}}
\newcommand{\vG}{\mathbf{G}}
\newcommand{\vH}{\mathbf{H}}
\newcommand{\vI}{\mathbf{I}}
\newcommand{\vJ}{\mathbf{J}}
\newcommand{\vK}{\mathbf{K}}
\newcommand{\vL}{\mathbf{L}}
\newcommand{\vM}{\mathbf{M}}
\newcommand{\vN}{\mathbf{N}}
\newcommand{\vO}{\mathbf{O}}
\newcommand{\vP}{\mathbf{P}}
\newcommand{\vQ}{\mathbf{Q}}
\newcommand{\vR}{\mathbf{R}}
\newcommand{\vS}{\mathbf{S}}
\newcommand{\vT}{\mathbf{T}}
\newcommand{\vU}{\mathbf{U}}
\newcommand{\vV}{\mathbf{V}}
\newcommand{\vW}{\mathbf{W}}
\newcommand{\vX}{\mathbf{X}}
\newcommand{\vY}{\mathbf{Y}}
\newcommand{\vZ}{\mathbf{Z}}

\newcommand{\va}{\mathbf{a}}
\newcommand{\vb}{\mathbf{b}}
\newcommand{\vc}{\mathbf{c}}
\newcommand{\vd}{\mathbf{d}}
\newcommand{\ve}{\mathbf{e}}
\newcommand{\vf}{\mathbf{f}}
\newcommand{\vg}{\mathbf{g}}
\newcommand{\vh}{\mathbf{h}}
\newcommand{\vi}{\mathbf{i}}
\newcommand{\vj}{\mathbf{j}}
\newcommand{\vk}{\mathbf{k}}
\newcommand{\vl}{\mathbf{l}}
\newcommand{\vm}{\mathbf{m}}
\newcommand{\vn}{\mathbf{n}}
\newcommand{\vo}{\mathbf{o}}
\newcommand{\vp}{\mathbf{p}}
\newcommand{\vq}{\mathbf{q}}
\newcommand{\vr}{\mathbf{r}}
\newcommand{\Vs}{\mathbf{s}}
\newcommand{\vt}{\mathbf{t}}
\newcommand{\vu}{\mathbf{u}}
\newcommand{\vv}{\mathbf{v}}
\newcommand{\vw}{\mathbf{w}}
\newcommand{\vx}{\mathbf{x}}
\newcommand{\vy}{\mathbf{y}}
\newcommand{\vz}{\mathbf{z}}

\newcommand{\vone}{\mathbf{1}}
\newcommand{\vzero}{\mathbf{0}}

\newcommand{\valpha}{{\boldsymbol{\alpha}}}
\newcommand{\vbeta}{{\boldsymbol{\beta}}}
\newcommand{\vgamma}{{\boldsymbol{\gamma}}}
\newcommand{\vdelta}{{\boldsymbol{\delta}}}
\newcommand{\vepsilon}{{\boldsymbol{\epsilon}}}
\newcommand{\vzeta}{{\boldsymbol{\zeta}}}
\newcommand{\veta}{{\boldsymbol{\eta}}}
\newcommand{\vtheta}{{\boldsymbol{\theta}}}
\newcommand{\viota}{{\boldsymbol{\iota}}}
\newcommand{\vkappa}{{\boldsymbol{\kappa}}}
\newcommand{\vlambda}{{\boldsymbol{\lambda}}}
\newcommand{\vmu}{{\boldsymbol{\mu}}}
\newcommand{\vnu}{{\boldsymbol{\nu}}}
\newcommand{\vxi}{{\boldsymbol{\xi}}}
\newcommand{\vomikron}{{\boldsymbol{\omikron}}}
\newcommand{\vpi}{{\boldsymbol{\pi}}}
\newcommand{\vrho}{{\boldsymbol{\rho}}}
\newcommand{\vsigma}{{\boldsymbol{\sigma}}}
\newcommand{\vtau}{{\boldsymbol{\tau}}}
\newcommand{\vupsilon}{{\boldsymbol{\upsilon}}}
\newcommand{\vphi}{{\boldsymbol{\phi}}}
\newcommand{\vchi}{{\boldsymbol{\chi}}}
\newcommand{\vpsi}{{\boldsymbol{\psi}}}
\newcommand{\vomega}{{\boldsymbol{\omega}}}

\newcommand{\rLambda}{\mathrm{\Lambda}}
\newcommand{\rSigma}{\mathrm{\Sigma}}

\makeatletter
\DeclareRobustCommand\onedot{\futurelet\@let@token\@onedot}
\def\@onedot{\ifx\@let@token.\else.\null\fi\xspace}
\def\eg{\emph{e.g}\onedot} \def\Eg{\emph{E.g}\onedot}
\def\ie{\emph{i.e}\onedot} \def\Ie{\emph{I.e}\onedot}
\def\cf{\emph{cf}\onedot} \def\Cf{\emph{C.f}\onedot}
\def\etc{\emph{etc}\onedot} \def\vs{\emph{vs}\onedot}
\def\wrt{w.r.t\onedot} \def\dof{d.o.f\onedot}
\def\etal{\emph{et al}\onedot}
\makeatother

\newcommand{\bib}[1]{\alert{[??]}}
\newcommand{\REF}[1]{\alert{[??]}}

\maketitle

\begin{abstract}
Semi-supervised learning is becoming increasingly important because it can combine data carefully labeled by humans with abundant unlabeled data to train deep neural networks. Classic methods
on semi-supervised learning that have focused on transductive learning have not been fully exploited in the inductive framework followed by modern deep learning. The same holds for the manifold assumption---that similar examples should get the same prediction.
In this work, we employ a transductive label propagation method that is based on the manifold assumption to make predictions on the entire dataset and use these predictions to generate pseudo-labels for the unlabeled data and train a deep neural network.
At the core of the transductive method lies a nearest neighbor graph of the dataset that we create based on the embeddings of the same network.
Therefore our learning process iterates between these two steps. We improve performance on several datasets especially in the few labels regime and show that our work is complementary to current state of the art.
\end{abstract}

\vspace{-12pt}
\section{Introduction}
\label{sec:intro}

\begin{figure}[t]
\vspace{-5pt}
\centering
\includegraphics[width=0.8\columnwidth]{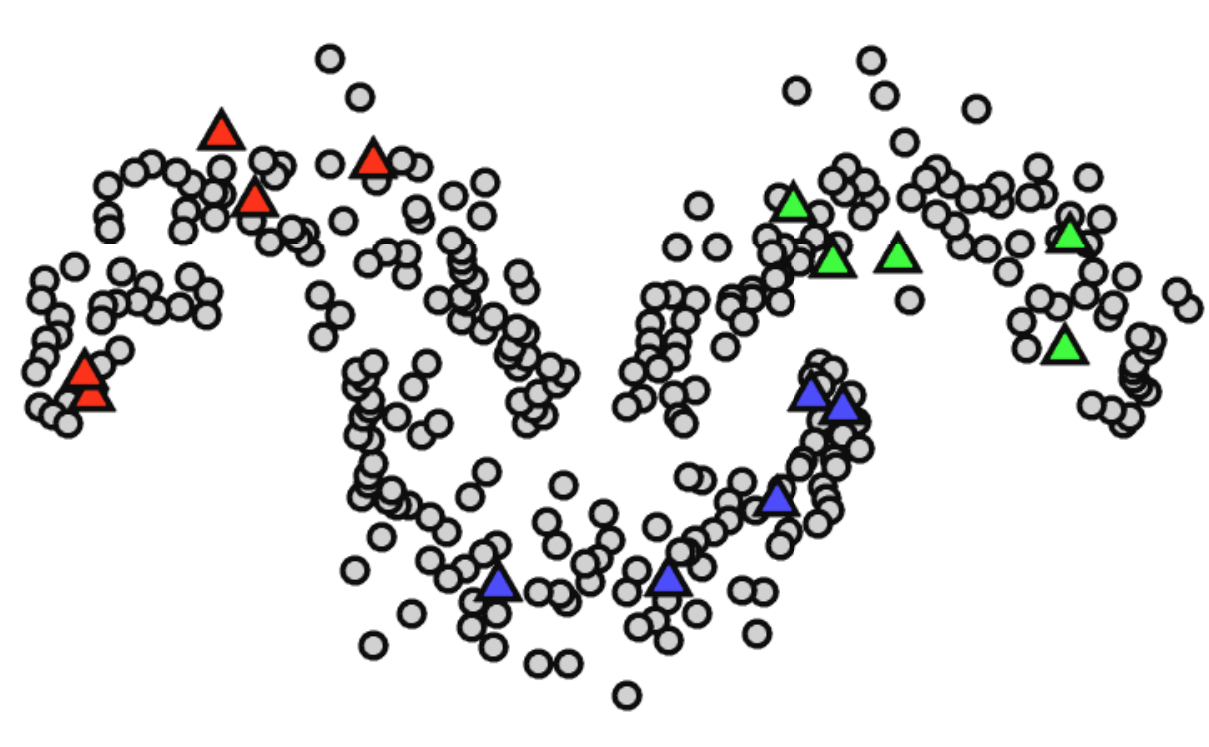}
\includegraphics[width=0.8\columnwidth]{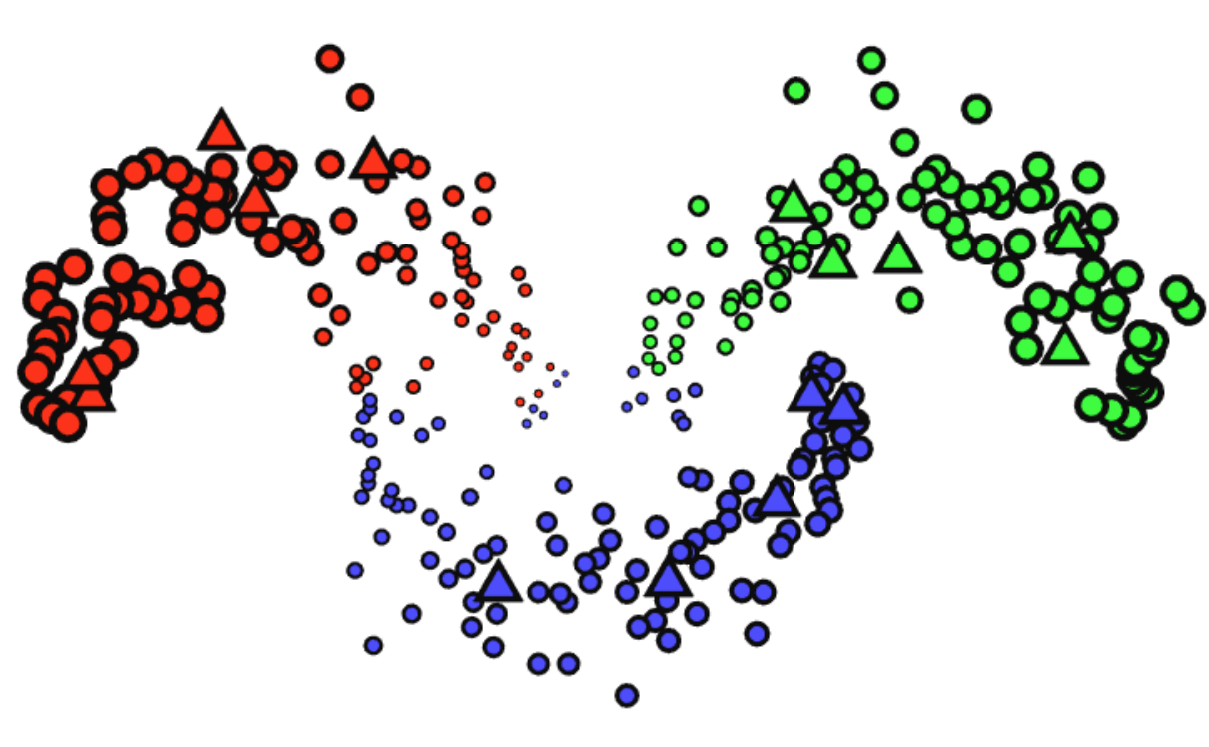}
\vspace{-8pt}
\caption{%
Label propagation on manifolds toy example. Triangles denote labeled, and circles un-labeled training data, respectively. Top: color-coded ground truth for labeled points, and gray color for unlabeled points. Bottom: color-coded pseudo-labels inferred by diffusion that are used to train the CNN. The size reflects the certainty of the pseudo-label prediction.
 \label{fig:intro}
 \vspace{-8pt}
 }
\end{figure}

Modern approaches to many computer vision problems exploit deep neural networks. These are popular for being very efficient and providing great performance at test time. The downside is a requirement of
large amounts of training examples, which are labeled either by humans or automatically on proxy tasks.

Visual data are available in large quantities, however, data reliably annotated by humans are still very scarce.
Obtaining large amounts of annotated training data for every single task is not only impractical, potentially costly, but it also turns out to be error prone. The low quality of crowd-sourced annotation is a common motivation to minimize the need of annotation.

In the domain of \emph{metric learning}, promising results have been recently achieved by \emph{unsupervised} methods for either learning from scratch or fine-tuning a supervised network for domain adaptation, which devise proxy tasks for learning. These tasks exploit the distribution of data in the original space, for instance pairwise relations of training examples
~\cite{wu2018unsupervised}, relations between examples and cluster centroids~\cite{caron2018deep}, or considering the manifold structure of data~\cite{ITA+18}. Alternatively, in \emph{self-supervised} learning, one can take advantage of additional information like spatial layout in images~\cite{DoGE15,GSK18} or temporal relation in videos~\cite{wang2017transitive,pathak2017learning}; or mine for such information in unstructured data by \emph{algorithmic supervision} using conventional methods~\cite{GARL17,RTC18}. However, most such proxy tasks are inferior when directly compared to laboriously annotated data by humans.

In classification, \emph{semi-supervised} methods attempt to reduce the number of labeled examples,
whereby the fully supervised performance on all data acts as an upper bound.
In \emph{transductive learning}~\cite{ZBL+03,ZhLG03},
label inference restricted to a given set of unlabeled examples is of interest.
In \emph{inductive learning}, the goal is generalization to new unseen data, while the original training data are discarded.
This is achieved \eg by combining classification loss on labeled data with unsupervised objectives on all data, where the latter act as regularization~\cite{WeRC08,TV17}. Or, an existing classifier can be used to assign pseudo-labels~\cite{Lee13,SJT16a}, which is another form of algorithmic supervision. Using a powerful classifier trained on carefully annotated data can provide high-quality pseudo-labels, opening the door to learning from real unlabeled, large scale data. In such \emph{omni-supervised} learning~\cite{RDG+18}, the fully supervised performance on the labeled part is actually the lower bound. This only refreshes the interest in inductive semi-supervised methods.

In this paper, we use efficient transductive
label propagation~\cite{ZBL+03} to infer pseudo-labels for unlabeled data, which are used to train the classifier.
Label propagation
is a graph-based method,
and in this work
the graph is constructed exploiting the embeddings obtained by the classification network itself.
Thus, the proposed method alternates between two steps. First, the network is trained from labeled and pseudo-labeled data. The second step uses the embeddings of the network trained in the previous step to construct a nearest neighbor graph. Label propagation
is then used to infer pseudo-labels for unlabeled images, as well as a certainty score per image and per class.
Training is performed on all data, using certainty-based weights.

We experimentally show on standard datasets that the proposed method outperforms other semi-supervised approaches. The less labeled data is available, the more pronounced the advantage of the proposed approach is.

\section{Related work}
\label{sec:related}
The literature is rich in the problem of \emph{semi-supervised learning} (SSL). The reader is advised to see~\cite{CSZ06} for an extensive overview.
The same holds for SSL in image classification~\cite{FWT09,GVS10,DG13,SSG12}.
In this section, we mostly restrict the discussion to approaches that use deep learning for SSL and perform the training on a large image collection with mini-batch optimization.

Prior work on semi-supervised deep learning for image classification is divided into two main categories.
The first consists of methods, \eg~\cite{GB05,LA17,SJT16b,TV17}, that add an \emph{unsupervised loss} term (often called a regularizer) into the loss function. This term is applied to either all images or only the unlabeled ones. Methods in the second category, \eg~\cite{Lee13,SGD+18}, assign \emph{pseudo-labels} to the unlabeled examples. The pseudo-labeled data are then used in training with a supervised loss, such as cross entropy.
Both categories use a standard loss term that is trained with supervision from labeled images.
A thorough evaluation of SSL deep image classification can be found in Miyato \etal~\cite{OOR+18}.

Our contribution belongs to the second category, and is conceptually and implementation-wise orthogonal to
the first. It is therefore straightforward to combine the proposed method with any method from the first category. We do combine it with~\cite{TV17} as shown in Section~\ref{sec:exp}.

\head{Unsupervised loss in deep SSL.}
Assuming that every training image, labeled or not, belongs to a single category, a natural requirement on the classifier is to make a confident prediction on the training set. This idea was formalized by Sajjadi~\etal~\cite{SJT16a}, where the regularizer is designed to minimize the entropy of the network output. Such a loss term is easily combined with other terms.
A similar combination is performed for denoising auto-encoders that are applied on all images in an unsupervised manner~\cite{RBH+15}.

A direction attracting a lot of attention is that of \emph{consistency loss}, where two related cases, \eg coming from two similar images, or made by two networks with related parameters, are encouraged to have similar network outputs. Sajjadi \etal~\cite{SJT16b} is the first, to our knowledge, to use a consistency loss between the outputs of a network on random perturbations of the same image.
Laine and Aila~\cite{LA17} rather apply consistency between the output of the current network and the temporal average of outputs during training.
The state-of-the-art \emph{mean teacher} (MT) method~\cite{TV17} replaces output averaging by averaging of network parameters.
Consistency loss is commonly measured by squared Euclidean distance.
The Jensen-Shannon divergence is used instead by Qiao \etal~\cite{QSZ18}, while complementarity of the two networks is enforced via adversarial examples.
A similar idea is proposed by Miyato \etal~\cite{MMI+18}.

\head{Pseudo-labeling in deep SSL.} 
Lee~\cite{Lee13} uses the current network to infer pseudo-labels of unlabeled examples, by choosing the most confident class. These pseudo-labels are treated like human-provided labels in the cross entropy loss. Its impact is similar to that of entropy minimization~\cite{SJT16a}; in both cases the network is forced to have more confident predictions.
The same principle is adopted by Shi \etal~\cite{SGD+18}, where the authors further add contrastive loss to the consistency loss.
Our method is different from all such prior work in that pseudo-labels are inferred by label propagation rather than network predictions.

\head{Label propagation} has been extensively used in a transductive setup (see chapter 11~\cite{CSZ06}).
Recently, Douze \etal~\cite{DSH+18} perform label propagation on a large image dataset with CNN descriptors for few shot learning.
Unseen images are classified via online label propagation, which requires storing the entire dataset, while the network is trained in advance and descriptors are fixed.
Our work is different in that we perform label propagation on the training set off-line while training the network, such that inference is possible without accessing the original training set.
\emph{Learning by association}~\cite{Haeusser_2017_CVPR} can been seen as two steps of propagation on a constrained bi-partite graph between labeled and unlabeled examples. \emph{Graph transduction game} (GTG)~\cite{EP12}, a form of label propagation, has been used for pseudo-labels~\cite{ETS18} as in our work, but in this case the network is pre-trained, the graph remains fixed and there is no weighting mechanism. We compare to this approach in Section~\ref{sec:exp}.

\section{Preliminaries}
\label{sec:pre}
In this section we formulate the \emph{semi-supervised learning} problem and then we discuss the classifier, different loss functions that are commonly used in prior work, and finally a transductive learning approach that our method is based on. In our experiments we use a \emph{convolutional neural network} (CNN) to perform image classification, but this formulation applies to any network architecture in any domain.

\head{Problem formulation.} We assume a collection of $n$ examples $X \defn (x_1, \ldots, x_l, x_{l+1}, \ldots, x_n)$ with $x_i \in \cX$. The first $l$ examples $x_i$ for $i \in L \defn\{1,\dots,l\}$, denoted by $X_L$, are labeled according to $Y_L \defn (y_1, \ldots, y_l)$ with $y_i \in C$, where $C \defn \{1,\dots,c\}$ is a discrete label set for $c$ classes. The remaining $u \defn n-l$ examples $x_i$ for $i \in U \defn\{l+1,\dots,n\}$, denoted by $X_U$, are unlabeled. The goal in SSL is to use all examples $X$ and labels $Y_L$ to train a classifier that maps previously unseen samples to class labels.

\head{Classifier.} The network takes an input example from $\cX$ and produces a vector of class confidence scores.
We denote it by $f\prm{\theta}: \cX \rightarrow \real^c$, where $\theta$ are the network parameters.
It is conceptually divided in two parts. 
The first is a feature extraction network $\phi\prm{\theta}: \cX \rightarrow \real^d$ mapping the input to a feature vector, or descriptor. We denote the descriptor of the $i$-th example by $\vv_i \defn \phi\prm{\theta}(x_i)$.
The second typically consists of a \emph{fully connected} (FC) layer applied on top of $\phi\prm{\theta}$ and followed by softmax, producing a vector of \emph{confidence scores}.
Function $f\prm{\theta}$ is the mapping from input space directly to confidence scores.
The output of the network for the $i$-th example is $f\prm{\theta}(x_i)$ and the \emph{prediction} is the one of maximum confidence score
\begin{equation}
\hat{y}_i \defn \arg\max_j f\prm{\theta}(x_i)_j,
\label{equ:netpredict}
\end{equation}
where subscript $j$ denotes the $j$-th dimension of the vector.

\head{Supervised loss.} In supervised learning, the network is trained by minimizing a \emph{supervised} loss term of the form
\begin{equation}
L_s(X_L, Y_L; \theta) \defn \sum_{i=1}^l \ell_s\left( f\prm{\theta}(x_i), y_i\right),
\label{eq:losuper}
\end{equation}
which applies only to labeled examples in $X_L$. Such term is part of the total loss when training a network in a semi-supervised setup~\cite{SGD+18,TV17,QSZ18}. A standard choice for the loss function $\ell_s$ in classification is \emph{cross-entropy}, given by $\ell_s(\Vs, y) \defn -\log \Vs_y$ for $\Vs \in \real^c$ and $y \in C$.

\head{Pseudo-labeling} is the process of assigning a pseudo-label $\hat{y}_i$ to each example $x_i$ for $i \in U$. Denoting by $\hat{Y}_U \defn (\hat{y}_{l+1}, \dots, \hat{y}_{n})$ the collection of pseudo-labels for $X_U$, the following additional \emph{pseudo-label} loss term applies
\begin{equation}
L_p(X_U, \hat{Y}_U; \theta) \defn \sum_{i=l+1}^n \ell_s\left( f\prm{\theta}(x_i), \hat{y}_i\right),
\label{eq:lopseudo}
\end{equation}
where again $\ell_s$ is any supervised loss function like cross-entropy.
An example is the approach proposed by Lee~\cite{Lee13}, who first train network $f\prm{\theta}$ with \eq{losuper} and then assign pseudo-labels according to (\ref{equ:netpredict}) for $i \in U$.

\head{Unsupervised loss} is another common alternative where the loss function applies to both labeled and unlabeled examples and encourages consistency under different transformations of the data or the network. The so-called \emph{consistency loss}~\cite{SGD+18,TV17,SGD+18} is defined as
\begin{equation}
L_u(X; \theta) \defn \sum_{i=1}^n \ell_u( f\prm{\theta}(x_i), f\prm{\tilde{\theta}}(\tilde{x}_i)),
\label{eq:lounsuper}
\end{equation}
where $\tilde{x}_i$ refers to a different transformation of example $x_i$.
Note that according to the standard practice of data augmentation, every forward pass of $x_i$ during training is performed under some random transformation.
Parameter set $\tilde{\theta}$ is either equal to $\theta$ or any other transformation of it, such as a moving average over the sequence of network updates~\cite{TV17}. A simple choice of $\ell_u$ is the squared Euclidean distance, \ie
$\ell_u( \Vs, \tilde{\Vs} ) \defn || \Vs - \tilde{\Vs}) ||^2$ for $\Vs, \tilde{\Vs} \in \real^c$,
forcing the two outputs to be as close as possible.

\head{Transductive learning} solves a more specific problem. Instead of training a generic classifier able to classify new, yet unseen, examples, the goal is to use $X$ and $Y_L$ to infer labels for examples in $X_U$.
In this work, we adopt the graph-based approach of Zhou \etal~\cite{ZBL+03} for transductive learning by diffusion\footnote{We first present the original approach and discuss our design choices in the following section.}.

\head{Diffusion for transductive learning~\cite{ZBL+03}.} Let $V = (\vv_1, \ldots, \vv_l, \vv_{l+1}, \ldots, \vv_{n})$ be the descriptor set, where $\vv_i$ corresponds to $x_i$ as defined earlier. A symmetric \emph{adjacency matrix} $W \in \real^{n \times n}$ with zero diagonal is constructed, whose elements $w_{ij}$ are non-negative pairwise similarities between $\vv_i$ and $\vv_j$. Its symmetrically normalized counterpart is given by $\cW = D^{-1/2} W D^{-1/2}$, where $D \defn \diag(W\vone_n)$ is the \emph{degree matrix} and $\vone_n$ is the all-ones $n$-vector. A $n \times c$ \emph{label matrix} $Y$ is defined with elements
\begin{align}
	Y_{ij} \defn \left\{
		\begin{array}{ll}
			1, & \mif i \in L \wedge y_i = j \\
			0, & \other.
	   \end{array}
	\right.
\vspace{-5pt}
\end{align}
That is, the rows of $Y$ corresponding to labeled examples are one-hot encoded labels and the rest are zero. Diffusion amounts to computing the $n \times c$ matrix
\begin{equation}
Z \defn (I-\alpha \cW)^{-1} Y,
\label{eq:zhou}
\vspace{-5pt}
\end{equation}
where $\alpha \in [0,1)$ is a parameter.
Finally, the class prediction for an unlabeled example $x_i$ is
\begin{equation}
\hat{y}_i \defn \arg\max_j z_{ij},
\label{eq:zhoupred}
\vspace{-5pt}
\end{equation}
where $z_{ij}$ is the $(i,j)$ element of matrix $Z$.

It is interesting to observe that matrix $Z$ as defined by~\eq{zhou} is the minimizer of the following quadratic cost function
\begin{equation}
J(Z) \defn	\frac{\alpha}{2} \hspace{-2pt} \sum_{i,j=1}^n \hspace{-3pt} w_{ij} \hspace{-2pt} \norm{ \frac{\vz_i}{\sqrt{d_{ii}}} - \frac{\vz_j}{\sqrt{d_{jj}}} }^2 \hspace{-4pt}+\hspace{-1pt} (1-\alpha) \norm{Y-Z}_F^2,
\label{eq:quad}
\end{equation}
where $\vz_i$ is the $i$-th row of matrix $Z$, $d_{ii}$ is the $i$-th diagonal diagonal element of $D$ and $\norm{\cdot}_F$ is the Frobenius norm. The first term encourages \emph{smoothness} such that nearby examples get the same predictions, while the second attempts to maintain predictions for the labeled examples~\cite{ZBL+03}.
\section{Method}
\label{sec:method}
\begin{figure*}
\vspace{-15pt}
\begin{center}
\begin{tabular}{cc}
\hspace{100pt}&
\input{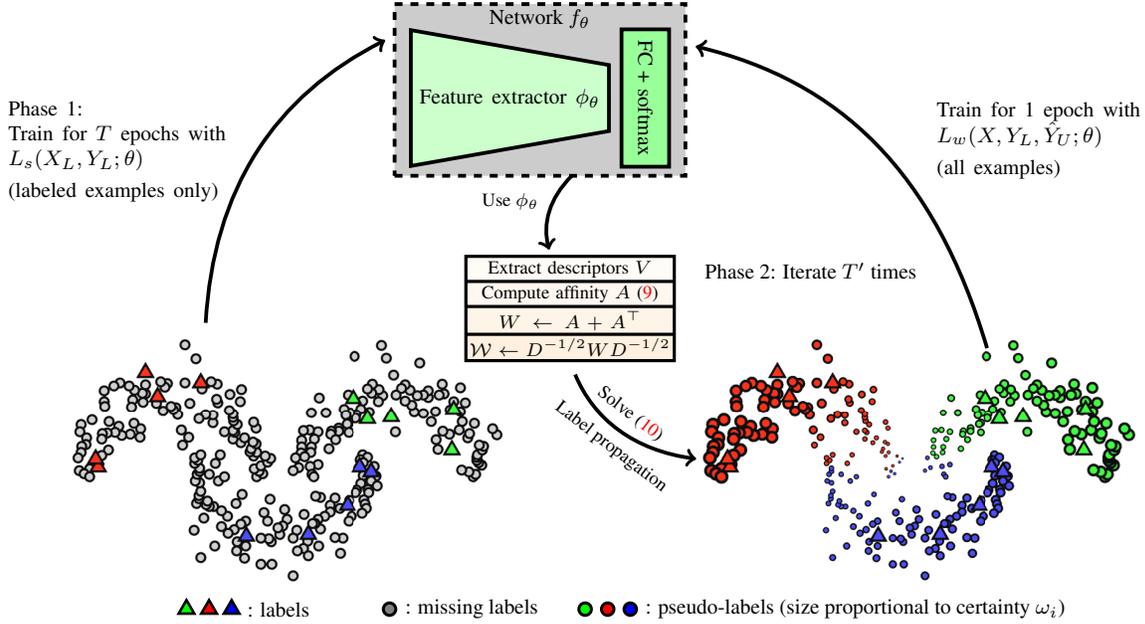}\\
\end{tabular}
\vspace{-8pt}
\caption{Overview of the proposed approach. Starting from a randomly initialized network, we first train it in a supervised fashion on the labeled examples. Then we initiate an iterative process where at each iteration we compute a nearest neighbor graph of the entire training set in the feature space of the current network, we propagate labels by transductive learning, and then we train the network on the entire training set, with true labels or pseudo-labels on the labeled or unlabeled examples respectively. The pseudo-labels are weighted per example and per class according to prediction certainty and inverse class population, respectively.
\label{fig:teaser}
\vspace{-8pt}
}
\end{center}
\end{figure*}

\begin{figure*}[t]
\centering
\begin{tabular}{ccc}
\includegraphics[width=0.31\textwidth]{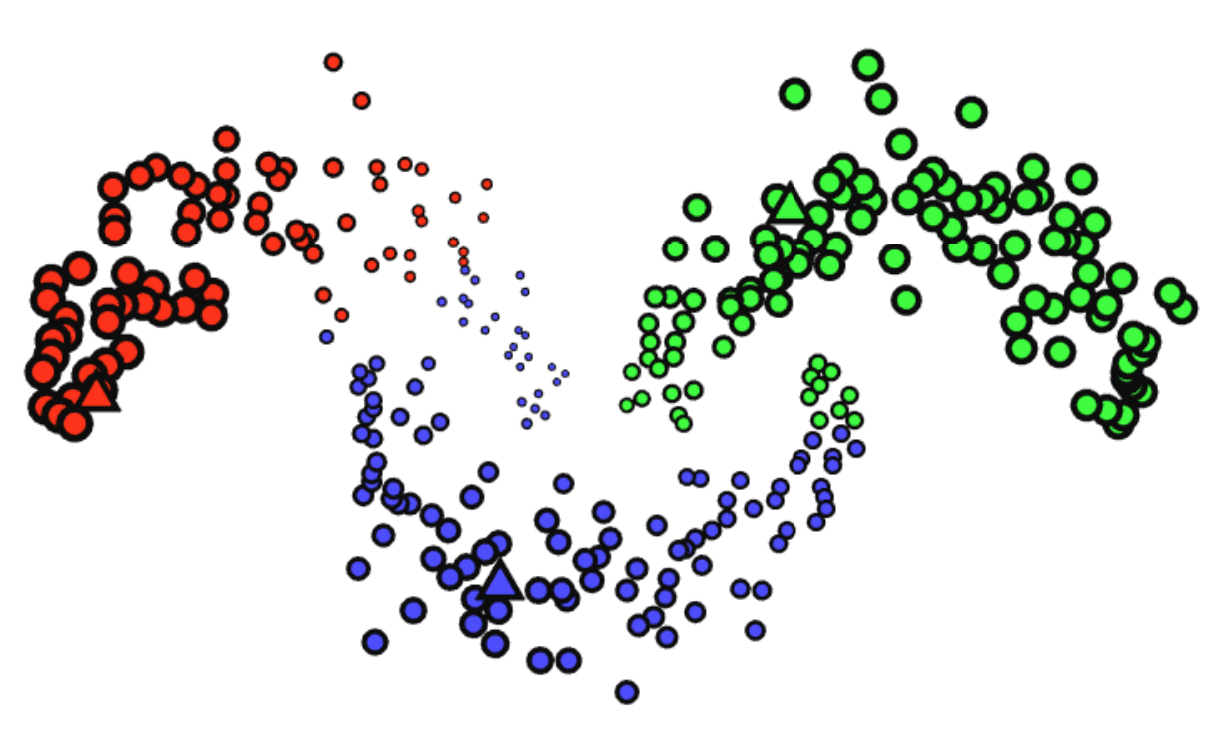}&
\includegraphics[width=0.31\textwidth]{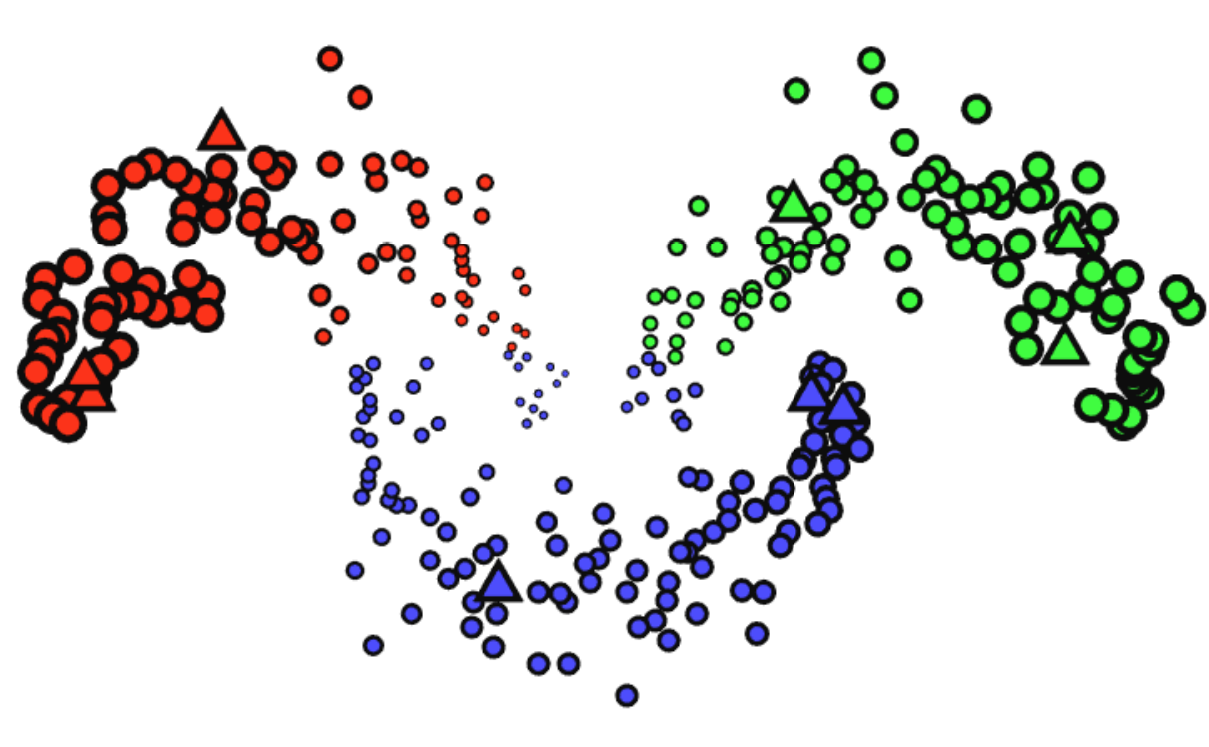}&
\includegraphics[width=0.31\textwidth]{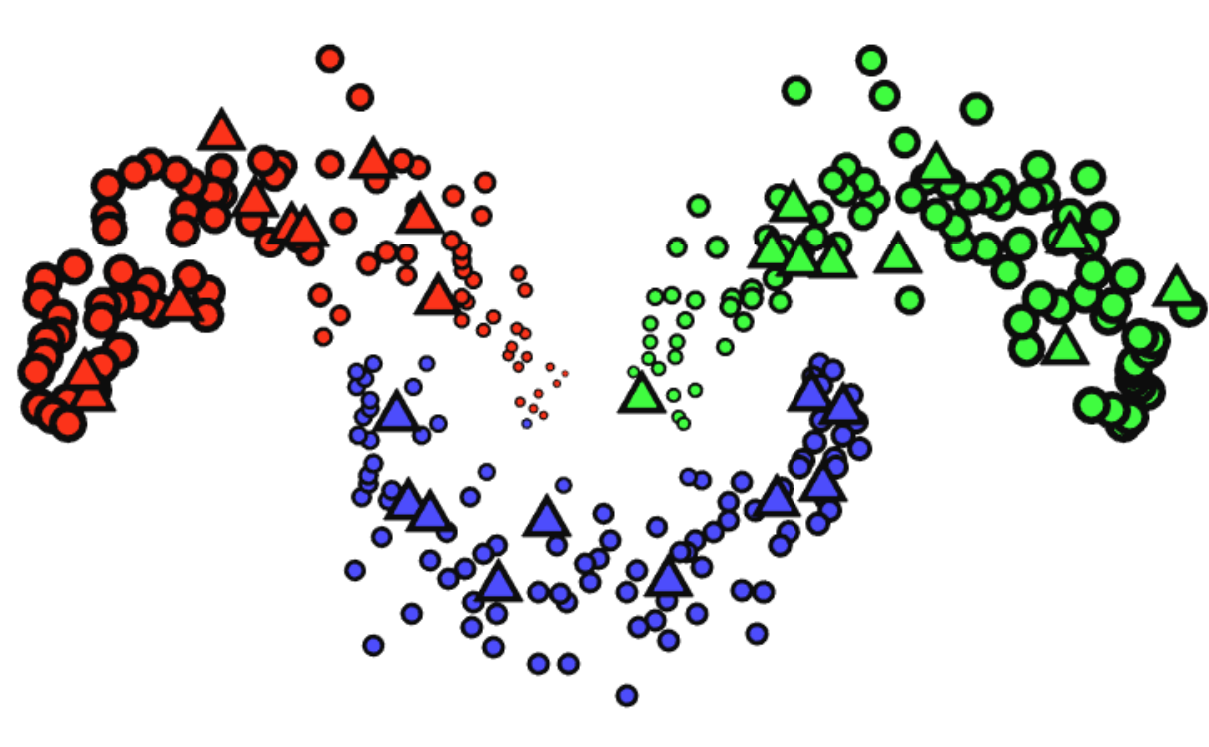} \\[-2pt]
1 labeled example & 3 labeled examples & 10 labeled examples \\[4pt]
\end{tabular}
\caption{Toy example with 300 examples demonstrating label propagation for different number of labeled examples. Triangle markers correspond the labeled examples and circles to the unlabeled ones which are finally pseudo-labeled by label propagation. The class is color-coded and the size of the circles corresponds to weight $\omega_i$. The true labels are the same as the example of Figure~\ref{fig:intro} (top).\label{fig:toy}
\vspace{-8pt}
}
\end{figure*}

In the following, we begin by providing an overview of our approach. We then develop the main elements of our solution, put everything together in a concrete algorithm, and discuss how our approach is complementary to approaches using unsupervised loss for SSL~\cite{TV17,SGD+18,SGD+18}. Finally, we discuss the relation to prior work that encourages smoothness in deep networks.

\head{Overview.} We introduce a new iterative process for semi-supervised learning that can be summarized as follows. First, we construct a nearest neighbor graph and perform label propagation by transductive learning on the training set. Then, we estimate of a weight reflecting the uncertainty of label propagation for each unlabeled example. Finally, we inject the obtained labels into the network training process. These ideas are developed below, while a graphical overview of the proposed approach is shown in Figure~\ref{fig:teaser}.

\head{Nearest neighbor graph.} Given a network with parameters $\theta$, we construct the descriptor set $V = (\vv_1, \ldots, \vv_l, \vv_{l+1}, \ldots, \vv_{n})$, where $\vv_i \defn \phi\prm{\theta}(x_i)$. A sparse \emph{affinity matrix} $A \in \real^{n \times n}$ with elements
\begin{equation}
	a_{ij} \defn
	\begin{cases}
		[\vv_i\tran \vv_j]_+^\gamma, & \mif i \ne j \wedge \vv_i \in \nn{k}(\vv_j)\\
		0, & \other
	\end{cases}
\label{eq:affinity}
\end{equation}
is constructed, where $\nn{k}$ denotes the set of $k$ nearest neighbors in $X$, and $\gamma$ is a parameter following recent work on manifold-based search~\cite{ITA+17}. Note that constructing the affinity matrix of the nearest neighbor graph is efficient even for large $n$~\cite{ITA+17}, while constructing the full affinity matrix as in Zhou \etal is not tractable. Then, let $W \defn A + A\tran$, which is indeed a symmetric nonnegative adjacency matrix with zero diagonal.

\head{Label propagation.} Estimating matrix $Z$ by (\ref{eq:zhou}) is impractical for large $n$ because the inverse matrix $(I-\alpha \cW)^{-1}$ is not sparse. We rather
use the the conjugate gradient (CG) method to
solve linear system
\begin{equation}
(I- \alpha \cW)Z = Y,
\label{eq:linsys}
\end{equation}
which applies because matrix $(I-\alpha \cW)$ is positive-definite. This solution is known to be faster than the iterative solution of Zhou \etal~\cite{ZBL+03}, and has been used in semi-supervised learning~\cite{ZhLR05}, interactive image segmentation~\cite{Grad06}, image retrieval~\cite{ITA+17} and semantic image segmentation~\cite{ChKo16}. Finally, we infer the pseudo-labels $\hat{Y}_U = (\hat{y}_{l+1}, \ldots, \hat{y}_{n} )$, where $\hat{y}_i$ is given by (\ref{eq:zhoupred}).

\head{Pseudo-label certainty and class balancing.} Inferring pseudo-labels from matrix $Z$ by hard assignment has two undesired effects: first, we define pseudo-labels on all unlabeled examples while clearly we do not have the same certainty for each example. Second, pseudo-labels may not be balanced over classes, which will impede learning.

To deal with the former issue we associate with each pseudo-label a weight reflecting the certainty of the prediction. We use \emph{entropy}, as a measure of uncertainty, to assign weight $\omega_i$ to example $x_i$, defined by
\begin{equation}
\omega_i \defn 1 - \frac{H(\hat{\vz}_i)} {\log (c)},
\label{eq:weight}
\end{equation}
where $\hat{Z}$ is the row-wise normalized counterpart of $Z$, \ie $\hat{z}_{ij} = z_{ij} / \sum_k{z_{ik}}$, and function $H: \real^{c}\rightarrow \real$ is the entropy function.
Weight $\omega_i$ is normalized in $[0,1]$ because $\log(c)$ is the maximum possible entropy in $\real^c$.

To deal with the latter issue of class imbalance, we assign weight $\zeta_j$ to class $j$ that is inversely proportional to class population, defined as $\zeta_j \defn (|L_j|+|U_j|)^{-1}$, where
$L_j$ (resp. $U_j$) are the examples labeled (resp. pseudo-labeled) as class $j$.

Given the above definitions of per-example and per-class weights, we associate the following \emph{weighted loss} to the labeled and pseudo-labeled examples
\begin{align}
L_w(X, Y_L, \hat{Y}_U; \theta) &\defn
	\sum_{i=1}^l \zeta_{y_i}
		\ell_s\left( f\prm{\theta}(x_i), y_i\right) \nonumber\\
		&+ 	\sum_{i=l+1}^n \omega_i \zeta_{\hat{y}_i}
		\ell_s\left( f\prm{\theta}(x_i), \hat{y}_i\right),
\label{eq:lopseudo2}
\end{align}
which is the sum of weighted versions of $L_s$~\eq{losuper} and $L_p$~\eq{lopseudo}.
In contrast to~\eq{lopseudo}, pseudo-labels originate in diffusion rather than network predictions.

A toy example showing the result of label propagation and the estimated weights is shown in Figure~\ref{fig:toy}.

\head{Iterative training.} Given the above definitions of nearest neighbor graph definition, label propagation, example/class weighting and pseudo-label loss, we plug those
components
into an iterative learning process. We begin by randomly initializing the network parameters $\theta$ and we train the network for $T$ epochs in a fully supervised manner on the $l$ labeled examples $X_L$ using the supervised loss term~(\ref{eq:losuper}). The trained network then provides the starting point for the following iterative process. First, we extract descriptors $V$ on the entire training set $X$ and compute nearest neighbors to construct the adjacency matrix $W$. Second, we perform label propagation by solving linear system~\eq{linsys} and assign pseudo-labels to unlabeled examples $X_U$ by~\eq{zhoupred}.
Finally, we train the network for one epoch on the entire training set $X$ using the weighted
loss
$L_w$~\eq{lopseudo2}.
We repeat this iterative process for $T'$ epochs. The above is summarized in Algorithm~\ref{alg:main}.

\begin{algorithm}
\caption{Label propagation for deep SSL\label{alg:main}}
\footnotesize
\algrenewcommand\algorithmicindent{0.5em}%
\begin{algorithmic}[1]
\Procedure{LPDSSL}{Training examples $X$, labels $Y_L$}
\State $\theta \gets$ initialize randomly
\For{$\text{epoch} \in [1,\ldots, T]$}
\State $\theta \gets$  \Call{Optimize}{$L_s(X_L, Y_L; \theta)$} \label{lin:train1} \comment{mini-batch optimization}
\EndFor
\For{$\text{epoch} \in [1,\ldots, T']$}
\State \algorithmicfor { $i \in \{1,\dots,n\}$} \algorithmicdo { $\vv_i \gets \phi\prm{\theta}(x_i)$} \comment{extract descriptors}
\State \algorithmicfor { $(i,j) \in \{1,\dots,n\}^2$} \algorithmicdo { $a_{ij} \gets$ affinity values (\ref{eq:affinity})} 
\State $W \gets A + A\tran$ \comment{symmetric affinity}
\State $\cW \gets D^{-1/2} W D^{-1/2}$ \comment{symmetrically normalized affinity}
\State $Z \gets$ solve (\ref{eq:linsys}) with CG \comment{diffusion}
\State \algorithmicfor { $(i,j) \in U \times C$} \algorithmicdo  { $\hat{z}_{ij} \gets {z}_{ij} / \sum_k {z}_{ik}$}\comment{normalize $Z$}
\State \algorithmicfor { $i \in U$} \algorithmicdo { $\hat{y}_i \gets \arg\max_j \hat{z}_{ij}$}\comment{pseudo-label}
\State \algorithmicfor { $i \in U$} \algorithmicdo { $\omega_i \gets$  certainty of $\hat{y}_i$ (\ref{eq:weight})}\comment{pseudo-label weight}
\State \algorithmicfor { $j \in C$} \algorithmicdo { $\zeta_j \gets (|L_j|+|U_j|)^{-1}$} \comment{class weight/balancing}
\State $\theta \gets$  \Call{Optimize}{$L_w(X, Y_L, \hat{Y}_U; \theta)$} \label{lin:train2} \comment{mini-batch optimization}
\EndFor
\EndProcedure
\end{algorithmic}

\end{algorithm}

Procedure $\Call{Optimize}$ refers to the mini-batch optimization of the corresponding loss term for one epoch, \ie all examples are fed to the network once. More details about batch construction are given in the implementation details.

\head{Combination with other approaches. }
Our contribution falls in the case of pseudo-label loss in the form of (\ref{eq:lopseudo}).
It is orthogonal to approaches that use unsupervised loss, for instance (\ref{eq:lounsuper}), applied to both labeled and unlabeled examples.
Combination of the two comes in a straightforward way by adding term (\ref{eq:lounsuper}) to the total loss optimized in lines \ref{lin:train1} and \ref{lin:train2} of Algorithm~\ref{alg:main}.
This is exactly the way we combine the proposed approach with the state-of-the-art Mean-Teacher approach~\cite{TV17} in our experiments.

\head{Discussion.} In an inductive framework, if $\vz_i / \sqrt{d_{ii}}$ is replaced by the network output $f_\theta(x_i)$ in the smoothness term of~\eq{quad}, then this becomes an unsupervised loss term, \eg like~\eq{lounsuper}, only now it encourages consistency between nearby example predictions. And indeed such solution is adopted \eg by Weston \etal~\cite{WeRC08}.
This is
not very efficient
because the adjacency matrix is typically sparse with non-zero-elements only on nearest neighbors, and then the gradient of the smoothness term will propagate from each example to its neighbors only at each iteration.

Our main idea therefore is that \emph{instead of just encouraging nearby examples to get the same predictions, we encourage all examples to get predictions same as the ones we would get by transductive learning} according to the quadratic cost~\eq{quad} and its solution $Z$~\eq{zhou}. Computing $Z$ is efficient because it is performed outside our main optimization process, \ie it does not need iterating on mini-batches of data and backpropagating through the network. Then, given $Z$, the main optimization process drives all examples directly to that solution, as if they were all labeled.

\section{Experiments}
\label{sec:exp}

We present the datasets used in our experiments and the SSL setup that is followed.
Then, we discuss the training details of our method and the methods reproduced for fair comparison.
Finally, we perform experiments to show the impact of different components involved in the proposed method and to compare with the state of the art.
All error rates reported are produced by our own implementation unless otherwise stated.
\subsection{Datasets}

We use three image classification datasets, namely CIFAR-10~\cite{KH09}, CIFAR-100~\cite{KH09} and Mini-ImageNet~\cite{VBL+16}.
Each dataset is used in an SSL setup where part of the training images are labeled and the rest are unlabeled.
We evaluate the performance on an independent test set.  Unless otherwise specified, error rate is reported in our experiments.

\head{CIFAR-10.}
The training set consists of 50k images coming from 10 classes, while the test set consists of 10k images from the same 10 classes.
All images have resolution $32\times32$.
Evaluation is performed with 50, 100, 200, and 400 labeled images per classes, corresponding to $l=500$, 1k, 2k, and 4k label images in total.
We use the same random selection of labeled images that is used in Mean Teacher~\cite{TV17} when available (1k, 2k and 4k labels).
The selection process is repeated $10$ times, resulting in 10 different dataset splits for SSL on CIFAR 10.
We follow the common practice which is to use each of them and report mean error and standard deviation.

\head{CIFAR-100.}
Similarly to CIFAR-10, CIFAR-100 has $50$k training and $10$k test images of resolution $32\times32$, coming from $100$ classes.
We follow a protocol equivalent to the one of CIFAR-10.
We evaluate with 40 and 100 labeled images per class, corresponding to 4k and 10k labeled  images in total.
There are $3$ such dataset splits, mean error and standard deviation are reported.

\head{Mini-ImageNet.}
We introduce an SSL evaluation setup for Mini-ImageNet~\cite{VBL+16} which is a subset of the well-known ImageNet~\cite{DSLLF09} dataset and has been previously used for few-shot learning~\cite{GK18}.
We use the train/test splits created in the work of Ravi and Larochelle~\cite{RL06}.
It consists of 100 classes with $600$ images per class, of resolution $84\times84$.
We randomly assign $500$ images from each class to the training set, and $100$ images to the test set.
The result is a train and test set of $50$k and $10$k images, respectively.
We create three dataset splits for the case of 40 and 100 labeled images per class that correspond to 4k and 10k labeled images in total.
Mean error and standard deviation over the three dataset splits are reported.

\subsection{Training}
We list the reproduced baselines, and provide training details per algorithm and dataset.

\textbf{Implementation.} We build our implementation on top of the publicly available Pytorch code for the Mean Teacher (MT) approach~\cite{TV17}\footnote{\scriptsize\url{https://github.com/CuriousAI/mean-teacher/tree/master/pytorch}}.
The fully supervised baseline and MT are reproduced identically as the original implementation.
In all our experiments SGD optimization is used.

\textbf{Networks.} Experiments on CIFAR-10 and CIFAR-100 are performed with the ``13-layer'' network that is used in prior work~\cite{LA17,TV17},
while on Mini-ImageNet, Resnet-18~\cite{HZRS16} is engaged.
Both networks consist of a feature extractor $\phi\prm{\theta}$ followed by an FC layer and softmax.
We add an $\l2$-normalization layer right after $\phi\prm{\theta}$ (before the FC layer) providing unit-norm descriptors for the graph construction. The same choice is also adopted in the fully supervised baseline. One exception is all variants of MT as we observed that the $\l2$-normalization layer slightly harms performance.
We normalize images to have channel-wise zero mean and unit variance over the entire training set.
Unlike prior work~\cite{TV17}, we do not normalize the input images with ZCA, nor add Gaussian noise to the input layer, which result in worse
performance according to our experiments.

\begin{table}
  \centering{\small
\begin{tabular}{@{\lsp}c@{\lsp}|@{\lsp}c@{\lsp}@{\lsp}c@{\lsp}|@{\lsp}c@{\lsp}}
Pseudo-labeling										& $\omega_i$ 		& $\zeta_j$		&  CIFAR-10 	\\
\hline
\multirow{4}{*}{Diffusion~\eqref{eq:zhoupred}}	& 				&     			&  $36.53\pm1.42$   		\\
												& 				& \OK			&  $36.17\pm1.98$			\\
												& \OK 			& 				&  $33.32\pm1.53$			\\
												& \OK 			& \OK			&  $\mathbf{32.40\pm1.80}$	\\ \hline
GTG~\cite{ETS18}					            & \OK 			& \OK			&  $35.20\pm2.23$			\\ \hline
Network~\eqref{equ:netpredict}					& \OK 			& \OK			&  $35.17\pm2.46$			\\
\hline
\end{tabular}
}
    \caption{Impact of weights $\omega_i$, class weights $\zeta_j$, and pseudo-labeling by diffusion prediction~\eqref{eq:zhoupred} or network prediction~\eqref{equ:netpredict}. Error rate is reported on CIFAR-10 with $500$ labels.
  \label{tab:abl}}
  \vspace{-2pt}
\end{table}

\begin{figure}
\centering
\input{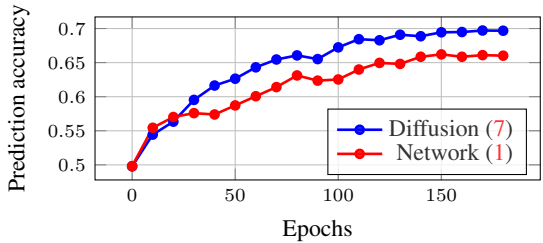}
\begin{tabular}{c}
{
\begin{tikzpicture}
\begin{axis}[%
	width=0.9\linewidth,
	height=0.45\linewidth,
	xlabel={\small Epochs},
	ylabel={\small Prediction accuracy},
	legend cell align={left},
	legend pos=south east,
    legend style={cells={anchor=east}, font =\small, fill opacity=0.8, row sep=-2.5pt},
    grid=both,
]
	\addplot[color=blue,     solid, mark=*,  mark size=1.5, line width=1.0] table[x=epochs, y expr={\thisrow{dfs}}] \predAcc;\leg{Diffusion~\eqref{eq:zhoupred}};
	\addplot[color=red,     solid, mark=*,  mark size=1.5, line width=1.0] table[x=epochs, y expr={\thisrow{cnn}}] \predAcc;\leg{Network~\eqref{equ:netpredict}};
\end{axis}
\end{tikzpicture}
}

\end{tabular}
\caption{
Accuracy of predicted pseudo-labels according to ground-truth
on CIFAR-10 with $500$ labeled images. Diffusion predictions~\eqref{eq:zhoupred} are compared against network predictions~\eqref{equ:netpredict}.
\label{fig:pseudoAcc}}
\vspace{-8pt}
\end{figure}

\begin{figure}
\vspace{-2pt}
\centering
\input{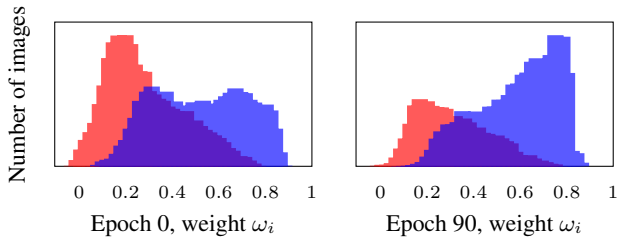}
\small
\begin{tabular}{@{\xssp}c@{\xssp}c@{\xssp}}
{
\begin{tikzpicture}
\begin{axis}[%
	width=0.6\linewidth,
	height=0.42\linewidth,
	ybar,
    xmax = 1,
    xtick style={draw=none},
    ytick style={draw=none},
    ymin = 0,
    grid=none,
    yticklabels={,,},
    xlabel={\small Epoch 0, weight $\omega_i$},%
    ylabel={\small Number of images},%
]
\addplot[color=red, fill = red,fill opacity=0.65,draw=none,bar width = 0.07] table[x expr={\thisrow{bins} - 0.025}, y expr={\thisrow{bad}}]  \weightHistInit;
\addplot[color=blue, fill = blue,fill opacity=0.65,draw=none,bar width = 0.07] table[x expr={\thisrow{bins} - 0.025}, y expr={\thisrow{ok}}]  \weightHistInit;
\end{axis}
\end{tikzpicture}
}
&
{
\begin{tikzpicture}
\begin{axis}[%
    width=0.6\linewidth,
    height=0.42\linewidth,
    ybar,
    xmax = 1,
    xtick style={draw=none},
    ytick style={draw=none},
    ymin = 0,
    grid=none,
    yticklabels={,,},
    xlabel={\small Epoch 90, weight $\omega_i$},%
]
\addplot[color=red, fill = red,fill opacity=0.65,draw=none,bar width = 0.07] table[x expr={\thisrow{bins} - 0.025}, y expr={\thisrow{bad}}]  \weightHistMiddle;
\addplot[color=blue, fill = blue,fill opacity=0.65,draw=none,bar width = 0.07] table[x expr={\thisrow{bins} - 0.025}, y expr={\thisrow{ok}}]  \weightHistMiddle;
\end{axis}
\end{tikzpicture}
}
\end{tabular}
\caption{
Distribution of weights $\omega_i$ for unlabeled images at epoch 0 (left) and epoch 90 (right) during the training of CIFAR-10 with $500$ labels.
Correct pseudo-labels according to ground-truth are shown in {\color{blue} blue} and incorrect in {\color{red} red}.
\label{fig:wDistr}
}
\vspace{-3pt}
\end{figure}

\begin{figure}
\centering
\input{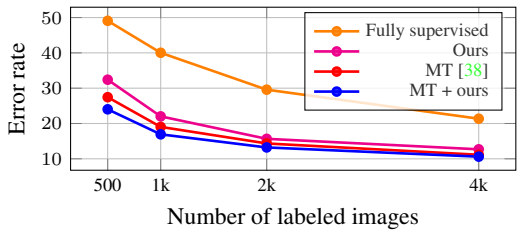}
\small
\begin{tabular}{c}
{
\begin{tikzpicture}
\begin{axis}[%
	width=0.9\linewidth,
	height=0.45\linewidth,
	xlabel={\small Number of labeled images},
	ylabel={\small Error rate},
	legend cell align={left},
	legend pos=north east,
    legend style={cells={anchor=east}, font =\scriptsize, fill opacity=0.8, row sep=-2.5pt},
   	xtick={500,1000,2000,4000},
   	xticklabels={500,1k,2k,4k},
    grid=both,
]
	\addplot[color=orange,     solid, mark=*,  mark size=1.5, line width=1.0] table[x=labeled, y expr={\thisrow{base}}] \labelAcc;\leg{Fully supervised};
	\addplot[color=magenta,     solid, mark=*,  mark size=1.5, line width=1.0] table[x=labeled, y expr={\thisrow{ours}}] \labelAcc;\leg{Ours};
	\addplot[color=red,     solid, mark=*,  mark size=1.5, line width=1.0] table[x=labeled, y expr={\thisrow{mt}}] \labelAcc;\leg{MT~\cite{TV17}};
	\addplot[color=blue,     solid, mark=*,  mark size=1.5, line width=1.0] table[x=labeled, y expr={\thisrow{ourmt}}] \labelAcc;\leg{MT + ours};

\end{axis}
\end{tikzpicture}
}

\end{tabular}
\caption{
Error rate versus number of labeled images on CIFAR-10 using different methods.
\label{fig:labelAcc}}
\end{figure}

\begin{figure}
\vspace{-6pt}
\newcommand{\imw}[1]{\includegraphics[height=1.3cm]{fig/ims/maxw/#1}}

\centering
\begin{narrow}
\begin{tabular}{cccccc}
\imw{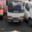} & 
\imw{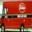} & 
\imw{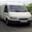} &
\imw{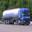} &
\imw{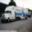} & 
\imw{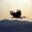}
\\[-4pt]
\footnotesize auto($0.82$) &
\footnotesize auto($0.82$) &
\footnotesize auto($0.82$) &
\footnotesize auto($0.82$) & 
\footnotesize auto($0.81$) &
\footnotesize ship($0.81$)
\\
\imw{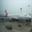} &
\imw{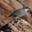} &
\imw{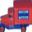} &
\imw{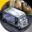} &
\imw{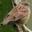} &
\imw{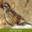}
\\[-4pt]
\footnotesize ship($0.81$) &
\footnotesize frog($0.80$) &
\footnotesize auto($0.80$) &
\footnotesize auto($0.80$) & 
\footnotesize frog($0.80$) &
\footnotesize frog($0.80$)


\end{tabular}
\end{narrow}


\caption{Examples of incorrectly pseudo-labeled images with highest $\omega_i$ in CIFAR-10. Predicted class and $\omega_i$ are shown below each image.\label{fig:omegaIms}}
\vspace{-3pt}
\end{figure}

\begin{table*}
\vspace{-8pt}
  \setlength\extrarowheight{-1pt}
\centering{\small
  \begin{tabular}{@{}lllll@{}}
	\toprule
	Dataset                                             & \multicolumn{4}{c}{CIFAR-10}  \\
	\cmidrule{2-5}
	Nb. labeled images						& 500				& 1000              & 2000              & 4000  \\
	\midrule
	Fully supervised							& $49.08\pm0.83$  	& $40.03\pm1.11$    & $29.58\pm0.93$    & $21.63\pm0.38$ \\
	\cmidrule{1-5}
	TDCNN~\cite{SGD+18}$^\dagger$			& -  				& $32.67\pm1.93$  	& $22.99\pm0.79$  	& $16.17\pm0.37$ \\

	Network prediction~\eqref{equ:netpredict} + weights								& $35.17\pm2.46$    & $23.79\pm1.31$    & $16.64\pm0.48$    & $13.21\pm0.61$ \\
	\textbf{Ours}: Diffusion prediction~\eqref{eq:zhoupred} + weights 							& $32.40\pm1.80$    & $22.02\pm0.88$    & $15.66\pm0.35$    & $12.69\pm0.29$ \\
	\cmidrule{1-5}
	VAT~\cite{MMI+18}$^\dagger$				& -       			& - 				& -    				& $11.36$ \\
	$\Pi$ model~\cite{LA17}$^\dagger$		& -       			& - 				& -    				& $12.36\pm0.31$ \\
	Temporal Ensemble~\cite{LA17}$^\dagger$	& -       			& - 				& -    				& $12.16\pm0.24$ \\
	MT~\cite{TV17}$^\dagger$				& -       			& $27.36\pm1.30$ 	& $15.73\pm0.31$   	& $12.31\pm0.28$ \\
	MT~\cite{TV17}							& $27.45\pm2.64$    & $19.04\pm0.51$    & $14.35\pm0.31$    & $11.41\pm0.25$ \\
	\textbf{MT + Ours}						& $\mathbf{24.02\pm2.44}$     & $\mathbf{16.93\pm0.70}$     & $\mathbf{13.22\pm0.29}$     & $\mathbf{10.61\pm0.28}$ \\
	\bottomrule
  \end{tabular}
}
  \caption{Comparison with the state of the art on CIFAR-10. Error rate is reported. ``13-layer'' network is used.
  The top part of the table corresponds to training with pseudo-labels, while the bottom part of the table includes methods that are complementary to ours, as shown by the combination of our method with MT. $\dagger$ denotes scores reported in prior work.
  \label{tab:cifar10}}
\end{table*}

\begin{table*}
  \setlength\extrarowheight{-1pt}
\centering{\small
  \begin{tabular}{@{}lllllll@{}}
	\toprule
	Dataset                                 & \multicolumn{2}{c}{CIFAR-100}			& \multicolumn{2}{c}{Mini-ImageNet-\textit{top1}}	& \multicolumn{2}{c}{Mini-ImageNet-\textit{top5}}	  \\
	\cmidrule{2-7}
	Nb. labeled images						& 4000				& 10000           						& 4000				& 10000				& 4000				& 10000   \\
	\midrule
	Fully supervised						& $55.43\pm0.11$  	& $40.67\pm0.49$    					& $74.78\pm0.33$			& $60.25\pm0.29$	& $53.07\pm0.68$		& $38.28\pm0.38$	\\
	\cmidrule{1-7}
	Ours									& $46.20\pm0.76$    & $38.43\pm1.88$    					& $\mathbf{70.29\pm0.81}$	& $57.58\pm1.47$	& $\mathbf{47.58\pm0.94}$	& $36.14\pm2.19$	\\
	MT~\cite{TV17}							& $45.36\pm0.49$    & $36.08\pm0.51$    					& $72.51\pm0.22$			& $57.55\pm1.11$	& $49.35\pm0.22$			& $32.51\pm1.31$	\\
	MT + Ours								& $\mathbf{43.73\pm0.20}$     & $\mathbf{35.92\pm0.47}$		& $72.78\pm0.15$			& $\mathbf{57.35\pm1.66}$	& $50.52\pm0.39$			& $\mathbf{31.99\pm0.55}$   \\
	\bottomrule
  \end{tabular}}
  \caption{Performance comparison on CIFAR-100 and Mini-ImageNet with 4k and 10k labeled images. Error rate is reported. ``13-layer'' network is used for CIFAR-100 and Resnet-18 is used for Mini-ImageNet. All methods are reproduced by us.
  \label{tab:others}}
\vspace{-8pt}
\end{table*}

\textbf{Hyper-parameters and training choices} are adapted from the MT method and implementation.
These are fixed for all approaches (re)produced by this work. The training is performed for 180 epochs in total.
Initial learning rate $l_0$ is decayed with cosine annealing~\cite{LH16} so that it would have reached zero after $210$ epochs, while $l_0 = 0.05$ on CIFAR-10, and $l_0 = 0.2$ on CIFAR-100 and Mini-ImageNet.
Random data augmentation is performed by $4\times 4$ random translations~\cite{TV17} followed by horizontal flip in CIFAR-10 and CIFAR-100.
On Mini-ImageNet, each image is randomly rotated by 10 degrees before random horizontal flip.
Batch size is 100 for CIFAR-10 and 128 for CIFAR-100 and Mini-ImageNet.
All other learning parameters remain unchanged from MT implementation.

\textbf{The fully supervised} approach corresponds to training with (\ref{eq:losuper}) and labeled images only.
MT uses the additional dual output trick with coefficient $0.01$.
Both these approaches are reproduced.

\textbf{Our approach} is performed with mini-batch size $B = B_U + B_L$, where $B_L$ images are labeled and $B_U$ images are originally unlabeled.
We set $B_L = 50$ for CIFAR-10 and $B_L=31$ for CIFAR100 and Mini-ImageNet.
Same is also applied for MT.
One epoch is defined as one pass through all originally unlabeled examples in the training set, meaning that images in $I_L$ appear multiple times per epoch.
We follow the same diffusion parameters as Iscen~\etal~\cite{ITA+17}.
We set $k=50$ for graph construction, $\gamma=3$ in~\eqref{eq:affinity}, and $\alpha = 0.99$ in~\eqref{eq:linsys}.
We solve~\eqref{eq:linsys} with at most $20$ iterations of CG.
Pairwise similarities for the graph are computed with the publicly available FAISS library~\cite{JDH17}.
Confidence weights $\omega_i$ are normalized over all examples s.t. $\max_i \omega_i = 1$.
Class weights $\zeta_j$ are normalized over $c$ classes such that the average class weight is $1$.
Pseudo-label predictions, $\omega_i$, and $\zeta_j$ are updated after each epoch.

To assess the benefit of diffusion, we finally evaluate a variant of our approach
where the pseudo-labels are not provided by diffusion but derived from the network with (\ref{equ:netpredict}) or from GTG propagation~\cite{ETS18} instead.
Training is performed with (\ref{eq:lopseudo2}), as with our method.
This
is in the spirit of pseudo-labeling in prior work~\cite{SGD+18,Lee13}.

\subsection{Ablation Study}
\label{sec:ablation}

We investigate the impact of different components of our method.
First, we study the effectiveness of weights introduced in the loss function~\eqref{eq:lopseudo2}.
Table~\ref{tab:abl} shows the classification performance on CIFAR-10 test set, when using only $500$ labeled examples for training and the rest of the training set is considered unlabeled.
Different weighting schemes are evaluated by setting all $\omega_i$ to one, all $\zeta_i$ to one, or both to one.
It is shown that both weights have positive contributions.
We also show the benefit of predicting with diffusion over predicting by the trained network or GTG propagation.
Pseudo-labeling by the network predictions uses examples that the network can already classify, while diffusion
allows for accurate predictions beyond those examples.
In Figure~\ref{fig:pseudoAcc}, we report the progress of the pseudo-label accuracy on unlabeled images $X_U$ throughout the training.
Diffusion predictions are consistently better than network predictions.

Figure~\ref{fig:wDistr} demonstrates how $\omega_i$ accurately estimates the certainty of the prediction.
From the plots we observe that predictions become more accurate as the training evolves, while at the beginning most examples are misclassified.
The proposed weighting mechanism is robust to
incorrect pseudo-labels and prevents
model collapse.
Figure~\ref{fig:omegaIms} shows some of the incorrectly pseudo-labeled images with high certainty $\omega_i$.
Most of the incorrect labels come from trucks labeled as automobiles or birds labeled as frogs.

\subsection{Comparison with the state-of-the-art}
We present a comparison with state-of-the-art on all 3 datasets in Tables~\ref{tab:cifar10} and~\ref{tab:others}.
The comparison includes performance reported in prior work and our reproduced results.
In the case of the work by Shi~\etal~\cite{SGD+18}, we only compare with their TDCNN variant which refers to pseudo-labeling for network training.
The other loss terms in their work are complementary to ours, similarly to MT.
We additionally compare with our implementation of pseudo-labeling with network predictions combined with the proposed weights.

The proposed approach performs the best out of the pseudo-label based approaches on CIFAR-10.
Results in Figure~\ref{fig:labelAcc} show that our benefit is larger when the number of labels is reduced.
The results on CIFAR-10 show that our approach is complementary to unsupervised loss, such as the one used by MT.
This combination achieves the best performance on this dataset.
The same holds for CIFAR-100 and Mini-ImageNet for 10k available labels.
Our method also achieves a lower error rate than temporal ensemble ($38.65 \pm 0.51$) and $\Pi$-model ($39.19 \pm 0.36$) on CIFAR-100~\cite{LA17} with $10$k labels.
On Mini-ImageNet with 4k available labels, the best performance is achieved when using our method without combining with Mean Teacher.

\section{Conclusions}
Most recent approaches for deep SSL rely on training with unsupervised loss on both labeled and unlabeled images.
We have proposed an approach that relies on graph-based label propagation to infer pseudo-labels for the unlabeled images.
An additional training set is formed with these pseudo-labels, which are shown to be more valuable than the pseudo-labels inferred by
the network itself. Our method is in principle complementary to unsupervised loss terms, which is
experimentally shown in this work.

\small{
\head{Acknowledgments}
This work is supported by the GA\v{C}R grant 19-23165S and the OP VVV funded project CZ.02.1.01/0.0/0.0/16\_019/0000765 ``Research Center for Informatics''.
}

{\small
\bibliographystyle{ieee}
\bibliography{egbib}

\begin{thebibliography}{10}\itemsep=-1pt

\bibitem{caron2018deep}
Mathilde Caron, Piotr Bojanowski, Armand Joulin, and Matthijs Douze.
\newblock Deep clustering for unsupervised learning of visual features.
\newblock {\em ECCV}, 2018.

\bibitem{ChKo16}
Siddhartha Chandra and Iasonas Kokkinos.
\newblock Fast, exact and multi-scale inference for semantic image segmentation
  with deep {Gaussian CRFs}.
\newblock In {\em ECCV}, 2016.

\bibitem{CSZ06}
Olivier Chapelle, Bernhard Scholkopf, and Alexander Zien.
\newblock {\em Semi-Supervised Learning}.
\newblock MIT Press, 2006.

\bibitem{DG13}
Dengxin Dai and Luc Van~Gool.
\newblock Ensemble projection for semi-supervised image classification.
\newblock In {\em ICCV}, 2013.

\bibitem{DoGE15}
Carl Doersch, Abhinav Gupta, and Alexei~A. Efros.
\newblock Unsupervised visual representation learning by context prediction.
\newblock In {\em ICCV}, 2015.

\bibitem{DSLLF09}
Wei Dong, Richard Socher, Li Li-Jia, Kai Li, and Li Fei-Fei.
\newblock Imagenet: A large-scale hierarchical image database.
\newblock In {\em CVPR}, June 2009.

\bibitem{DSH+18}
Matthijs Douze, Arthur Szlam, Bharath Hariharan, and Herv{\'e} J{\'e}gou.
\newblock Low-shot learning with large-scale diffusion.
\newblock In {\em CVPR}, 2018.

\bibitem{ETS18}
Ismail Elezi, Alessandro Torcinovich, Sebastiano Vascon, and Marcello Pelillo.
\newblock Transductive label augmentation for improved deep network learning.
\newblock {\em arXiv preprint arXiv:1805.10546}, 2018.

\bibitem{EP12}
Aykut Erdem and Marcello Pelillo.
\newblock Graph transduction as a noncooperative game.
\newblock {\em Neural Computation}, 24, 2012.

\bibitem{FWT09}
Rob Fergus, Yair Weiss, and Antonio Torralba.
\newblock Semi-supervised learning in gigantic image collections.
\newblock In {\em NIPS}, 2009.

\bibitem{GK18}
Spyros Gidaris and Nikos Komodakis.
\newblock Dynamic few-shot visual learning without forgetting.
\newblock In {\em CVPR}, 2018.

\bibitem{GSK18}
Spyros Gidaris, Praveer Singh, and Nikos Komodakis.
\newblock Unsupervised representation learning by predicting image rotations.
\newblock In {\em ICLR}, 2018.

\bibitem{GARL17}
Albert Gordo, Jon Almazan, Jerome Revaud, and Diane Larlus.
\newblock End-to-end learning of deep visual representations for image
  retrieval.
\newblock {\em IJCV}, 124(2), 2017.

\bibitem{Grad06}
Leo Grady.
\newblock Random walks for image segmentation.
\newblock {\em IEEE Trans. PAMI}, 28(11):1768--1783, 2006.

\bibitem{GB05}
Yves Grandvalet and Yoshua Bengio.
\newblock Semi-supervised learning by entropy minimization.
\newblock In {\em NIPS}, 2005.

\bibitem{GVS10}
Matthieu Guillaumin, Jakob Verbeek, and Cordelia Schmid.
\newblock Multimodal semi-supervised learning for image classification.
\newblock In {\em CVPR}, 2010.

\bibitem{Haeusser_2017_CVPR}
Philip Haeusser, Alexander Mordvintsev, and Daniel Cremers.
\newblock Learning by association -- a versatile semi-supervised training
  method for neural networks.
\newblock In {\em CVPR}, 2017.

\bibitem{HZRS16}
Kaiming He, Xiangyu Zhang, Shaoqing Ren, and Jian Sun.
\newblock Deep residual learning for image recognition.
\newblock In {\em CVPR}, 2016.

\bibitem{ITA+18}
Ahmet Iscen, Giorgos Tolias, Yannis Avrithis, and Ondrej Chum.
\newblock Mining on manifolds: Metric learning without labels.
\newblock In {\em CVPR}, 2018.

\bibitem{ITA+17}
Ahmet Iscen, Giorgos Tolias, Yannis Avrithis, Teddy Furon, and Ondrej Chum.
\newblock Efficient diffusion on region manifolds: Recovering small objects
  with compact cnn representations.
\newblock In {\em CVPR}, 2017.

\bibitem{JDH17}
Jeff Johnson, Matthijs Douze, and Herv{\'e} J{\'e}gou.
\newblock Billion-scale similarity search with gpus.
\newblock {\em arXiv preprint arXiv:1702.08734}, 2017.

\bibitem{KH09}
Alex Krizhevsky and Geoffrey Hinton.
\newblock Learning multiple layers of features from tiny images.
\newblock Technical report, University of Toronto, 2009.

\bibitem{LA17}
Samuli Laine and Timo Aila.
\newblock Temporal ensembling for semi-supervised learning.
\newblock In {\em ICLR}, 2017.

\bibitem{Lee13}
Dong-Hyun Lee.
\newblock Pseudo-label: The simple and efficient semi-supervised learning
  method for deep neural networks.
\newblock In {\em ICMLW}, 2013.

\bibitem{LH16}
Ilya Loshchilov and Frank Hutter.
\newblock Sgdr: Stochastic gradient descent with warm restarts.
\newblock {\em ICLR}, 2017.

\bibitem{MMI+18}
Takeru Miyato, Shin-ichi Maeda, Shin Ishii, and Masanori Koyama.
\newblock Virtual adversarial training: a regularization method for supervised
  and semi-supervised learning.
\newblock {\em IEEE Trans. PAMI}, 2018.

\bibitem{OOR+18}
Avital Oliver, Augustus Odena, Colin Raffel, Ekin~D Cubuk, and Ian~J
  Goodfellow.
\newblock Realistic evaluation of deep semi-supervised learning algorithms.
\newblock In {\em ICLRW}, 2018.

\bibitem{pathak2017learning}
Deepak Pathak, Ross~B Girshick, Piotr Doll{\'a}r, Trevor Darrell, and Bharath
  Hariharan.
\newblock Learning features by watching objects move.
\newblock In {\em CVPR}, 2017.

\bibitem{QSZ18}
Siyuan Qiao, Wei Shen, Zhishuai Zhang, Bo Wang, and Alan Yuille.
\newblock Deep co-training for semi-supervised image recognition.
\newblock In {\em ECCV}, 2018.

\bibitem{RTC18}
Filip Radenovi{\'c}, Giorgos Tolias, and Ond{\v{r}}ej Chum.
\newblock Fine-tuning {CNN} image retrieval with no human annotation.
\newblock {\em IEEE Trans. PAMI}, 2018.

\bibitem{RDG+18}
Ilija Radosavovic, Piotr Dollar, Ross Girshick, Georgia Gkioxari, and Kaiming
  He.
\newblock Data distillation: Towards omni-supervised learning.
\newblock In {\em CVPR}, 2018.

\bibitem{RBH+15}
Antti Rasmus, Mathias Berglund, Mikko Honkala, Harri Valpola, and Tapani Raiko.
\newblock Semi-supervised learning with ladder networks.
\newblock In {\em NIPS}, 2015.

\bibitem{RL06}
Sachin Ravi and Hugo Larochelle.
\newblock Optimization as a model for few-shot learning.
\newblock In {\em ICLR}, 2016.

\bibitem{SJT16b}
Mehdi Sajjadi, Mehran Javanmardi, and Tolga Tasdizen.
\newblock Mutual exclusivity loss for semi-supervised deep learning.
\newblock In {\em ICIP}, 2016.

\bibitem{SJT16a}
Mehdi Sajjadi, Mehran Javanmardi, and Tolga Tasdizen.
\newblock Regularization with stochastic transformations and perturbations for
  deep semi-supervised learning.
\newblock In {\em NIPS}, 2016.

\bibitem{SGD+18}
Weiwei Shi, Yihong Gong, Chris Ding, Zhiheng Ma, Xiaoyu Tao, and Nanning Zheng.
\newblock Transductive semi-supervised deep learning using min-max features.
\newblock In {\em ECCV}, 2018.

\bibitem{SSG12}
Abhinav Shrivastava, Saurabh Singh, and Abhinav Gupta.
\newblock Constrained semi-supervised learning using attributes and comparative
  attributes.
\newblock In {\em ECCV}, 2012.

\bibitem{TV17}
Antti Tarvainen and Harri Valpola.
\newblock Mean teachers are better role models: Weight-averaged consistency
  targets improve semi-supervised deep learning results.
\newblock In {\em NIPS}, 2017.

\bibitem{VBL+16}
Oriol Vinyals, Charles Blundell, Tim Lillicrap, Daan Wierstra, et~al.
\newblock Matching networks for one shot learning.
\newblock In {\em NIPS}, 2016.

\bibitem{wang2017transitive}
Xiaolong Wang, Kaiming He, and Abhinav Gupta.
\newblock Transitive invariance for selfsupervised visual representation
  learning.
\newblock In {\em ICCV}, 2017.

\bibitem{WeRC08}
Jason Weston, Fr\'ed\'eric Ratle, and Ronan Collobert.
\newblock Deep learning via semi-supervised embedding.
\newblock In {\em ICML}, 2008.

\bibitem{wu2018unsupervised}
Zhirong Wu, Yuanjun Xiong, Stella Yu, and Dahua Lin.
\newblock Unsupervised feature learning via non-parametric instance-level
  discrimination.
\newblock {\em CVPR}, 2018.

\bibitem{ZBL+03}
Dengyong Zhou, Olivier Bousquet, Thomas~Navin Lal, Jason Weston, and Bernhard
  Sch{\"o}lkopf.
\newblock Learning with local and global consistency.
\newblock In {\em NIPS}, 2003.

\bibitem{ZhLR05}
Xiaojin Zhu, John Lafferty, and Ronald Rosenfeld.
\newblock {\em Semi-Supervised Learning with Graphs}.
\newblock PhD thesis, Carnegie Mellon University, Language Technologies
  Institute, School of Computer Science Pittsburgh, PA, 2005.

\bibitem{ZhLG03}
Xiaojin Zhu, John~D Lafferty, and Zoubin Ghahramani.
\newblock Semi-supervised learning: From {Gaussian} fields to {Gaussian}
  processes.
\newblock Technical report, 2003.

\end{thebibliography}
}
\end{document}